# Vision-model-based Real-time Localization of Unmanned Aerial Vehicle for Autonomous Structure Inspection under GPS-denied Environment

**Zhexiong Shang,[1] and Zhigang Shen, Ph.D.[2]**

[1]NH 113, Durham School of Architectural Engineering & Construction, University of Nebraska-Lincoln, Lincoln, NE, 68588; e-mail: szx0112@huskers.unl.edu
[2]NH 113, Durham School of Architectural Engineering & Construction, University of Nebraska-Lincoln, Lincoln, NE, 68588; e-mail: shen@unl.edu

**ABSTRACT**

UAVs have been widely used in visual inspections of buildings, bridges and other structures. In either outdoor autonomous or semi-autonomous flights missions strong GPS signal is vital for UAV to locate its own positions. However, strong GPS signal is not always available, and it can degrade or fully loss underneath large structures or close to power lines, which can cause serious control issues or even UAV crashes. Such limitations highly restricted the applications of UAV as a routine inspection tool in various domains. In this paper a vision-model-based real-time self-positioning method is proposed to support autonomous aerial inspection without the need of GPS support. Compared to other localization methods that requires additional onboard sensors, the proposed method uses a single camera to continuously estimate the inflight poses of UAV. Each step of the proposed method is discussed in detail, and its performance is tested through an indoor test case.

**INTRODUCTION**

Many recent studies investigated the performance of UAV on various inspection applications, such as building inspection (Eschmann et al. 2012), wind turbine monitoring (Schäfer et al. 2016), construction site reconstruction (Shang and Shen 2018) and bridge condition analysis (Metni and Hamel 2007). In these applications, the flight paths of UAVs are pre-determined based on the geometry of the inspection structure and the sensing requirements. Each path is a set of GPS waypoints that is designed either manually or through specific path planning algorithms. These waypoints are then used to guide the inflight positions of UAVs for aerial images collection. Most commercially available UAVs rely on the GNSS navigation system to enable the autonomous waypoint following. More advanced flight control systems use the sensor fusion techniques to provide safe and smooth flight by integrating the satellite position information from GPS with the acceleration and angular rate measurements from the Inertial Measurement Units (IMU). These widely applied navigation systems support the UAV-based autonomous inspection where the GPS signal is strong. However, strong GPS signal is not always available, and it can degrade or fully loss underneath large structures or close to power lines, which can cause serious control issues or even UAV crashes. Such limitations highly restrict the applications of UAVs as a routine inspection tool in many aspects. Therefore, approaches that can locate the inflight positions of UAV under GPS-denied environment is high demanded.

Among the different localization techniques, the vision-based approaches have showed the great success on tracking and locating moving objects in clutter environments. Unlike other sensors that requires additional payloads on UAV, the vision-based approaches can use the onboard camera for UAV real-time localization. The existing visual localization techniques are categorized into two major groups: the model-free method and the model-based method. The mode-free method, such as Simultaneously Localization and Mapping (SLAM), is mostly used in exploring unstructured environments. The method incrementally builds a live map of the surroundings using monocular or stereo cameras, and at the same time, sensing the locations of robot within the map (Kim 2012). This method uses the salient features detected in each frame to continuously estimate the camera locations. However, for civil structures, such as building walls and bridge decks, those salient features may not always available in the scene. In addition, SLAM only outputs a relative map, additional processing is still needed in order to locate the inflight positions of UAV in the world coordinates.

Compared to the model-free method, the model-based method assumes the model of the sensing environment is known. This make the model-based method efficient as only the visual information extracted from the model are utilized to estimate the camera locations. For inspection UAVs, the onboard camera is always focused on the inspected structures. Those structure centered images can be utilized by the model-based method to locate the UAV poses. Although the model-based method has been applied in tracking and localization applications in various domains, such as robot manipulation (Choi and Christensen 2010), autonomous ground and aerial robots (Manz et al. 2011; Teuliere et al. 2015) and augment reality (Lowney and Raj 2016; Wuest et al. 2005), few studies evaluated its usage on localizing the inflight positions of UAV for aerial structural inspection.

Therefore, to fill this knowledge gap, an improved model-based, real-time visual localization method is proposed in this study. As the first step towards the autonomous UAV inspection under GPS-denied environment, the proposed method aims to continuously estimate the 6d poses of UAV through the onboard camera without known its flight trajectory. Each step of the proposed method is discussed in detail, and the practicability of the method is verified through an indoor environment test case.

**METHODOLODY**

The workflow of the proposed method is shown in Fig. 1. The proposed method is an edge-based visual localization algorithm developed based on the RAPID tracker (Harris and Stennett 1990). Each step of the proposed method is explained below.

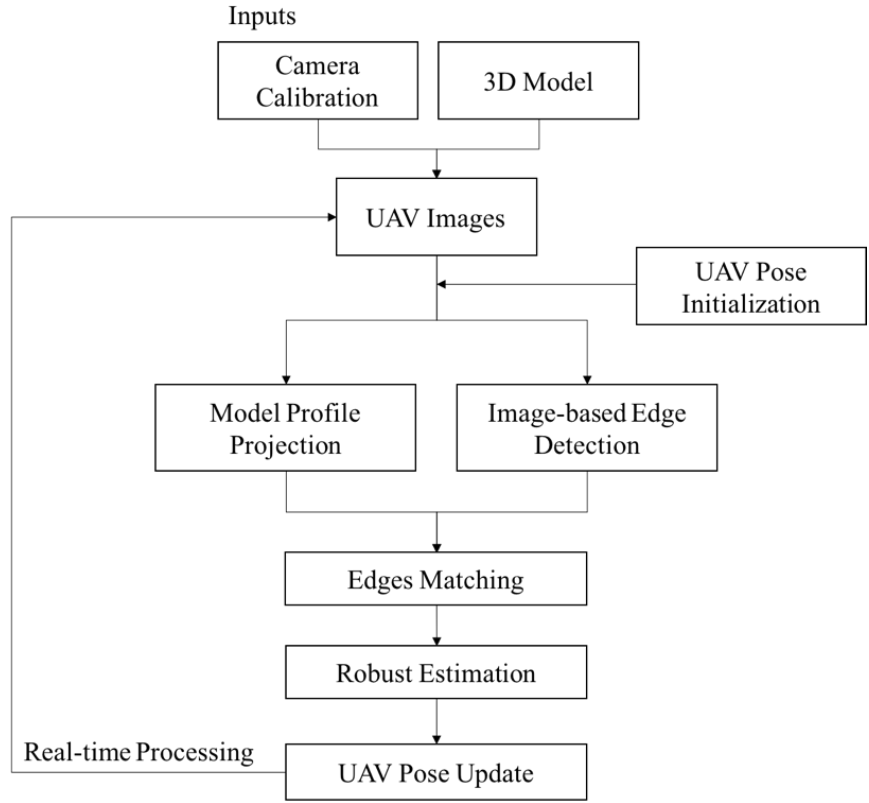

**Figure 1** Workflow of the proposed method

**Inputs**
In this method, we assume the 3D geometrical model of the inspection structure is known. This 3D model is either converted from the design drawings, or through as-built 3D reconstruction. The method also requires the inspection camera is pre-calibrated in order to acquire the correct projection from 3D model of the structure to the 2D image planes. In this study, the camera is calibrated by solving the Perspective-n-Point (PnP) problem as in (Heikkila and Silven 1997).

**UAV Pose Initialization**
In this study, we assume the camera is directly attached on the UAV, thus, no relational movement is existed between UAV and onboard camera. The positions and orientations of UAV is computed by multiplying a transformation matrix from the camera poses. Due to the method follows the pipeline of the most tracking algorithms, an initial pose of the camera is needed. Such pose can be either defined manually or by starting the UAV at a known position. It is noted that both of these pose initialization strategies require the inspection target in the view in order for the accurate pose projection. Based on the initial pose, the 3D model is then re-projected onto the next image plane with the camera projection matrix (Haralick and Shapiro 1992).

**Model Profile Projection**
After the initial pose is determined, we compute the projected 3D model in the image plane. For 3D structures, it is vital to identify which surface of the structure is visible, and which one is occluded. Detecting and matching edges only at the visible surface can ensure the accuracy of pose estimation. In this study, the visible surfaces are identified by computing the dot product of

― 3 ―

the normal of each structural surface and the camera ray. If the sign of this dot product is negative, then the surface is visible, otherwise it is occluded. For each visible surface, edges connect to the surface are defined as visible edges.

**Image-based Edge Detection**
At this step, the edges information is extracted from each frame. Unlike other visual tracking methods where the typical edge detection algorithms, such as canny edge detector, are applied to extract the edge information. Our method implement the LSD line segment detector for edge detection (Von Gioi et al. 2010). This modification is based on the condition that the geometries of the most man-made 3D structures are composed of straight lines or likely straight lines in images. Using this line detector can automatically omit the short/curve edges in the background (e.g. trees, cloud, birds and etc.) which reduces the computational workload and, at the same time, increases the accuracy of edge matching results.

**Edge Matching**
In order to match the projected profile of the 3D model and the edges detected in the 2D image plane, an appropriate edge-to-edge matching solution is demanded. In this study, we provide the similar strategy as in (Pupilli and Calway 2006) to match the orthogonal distance between two edges. The method first samples control points on each edge, then compare the perpendicular distances between the points of different edges. The sampling rate along each edge is proportional to the projected model size in image plane. The correspondences between the control points detected in the image plane and from the 3D projected model are measured, and the camera pose is updated by minimizing such correspondences. For control point distance measurement, we applied the one-dimensional search strategy in the RAPID tracker (Harris and Stennett 1990). This edge matching strategy iteratively measures all the correspondences between the 3D projection model and the 2D image plane.

**Robustness Estimation**
After the matched edges are measured, the next step is to find the best matches in order to provide robust camera pose estimation. In this study, we employed the low-level multiple hypotheses method (Teulière et al. 2010) to identify the best match in the view. The method computes the camera pose by selecting the matching with minimal distance errors. In this study, we identify these minimal matching using Random Sample Consensus (RANSAC). These refined matching results are then used to compute the fine pose of camera at each frame.

**UAV Pose Update**
After the fine matched edges are identified, the new camera pose can be updated by computing the transformation matrix between the projected model profile in previous frame and the image detected edges in the the current frame. RANSAC is used as the estimator to eliminate the outliners and refine the match results. These refined matching are then used to estimate the new camera poses with the classic PnP solution. This computed new camera pose is then used to estimate the UAV inflight position through an inverse transformation.

**EXPERIMENTAL SETUP**

In this section, we test the performance the proposed method on an indoor UAV platform. A single rectangular box is selected as inspection target of the test case. The selected UAV is an indoor autonomous inspection bundle developed in the authors' prior work (Shang and Shen 2018). The platform is composed of a micro quadrotor (i.e. Crazyflie 2.0), a FPV camera, a flow deck, a wireless video receiver, a Xbox controller and a PC laptop. The aerial platform, hardware configuration and the test environment are shown in Fig. 2.

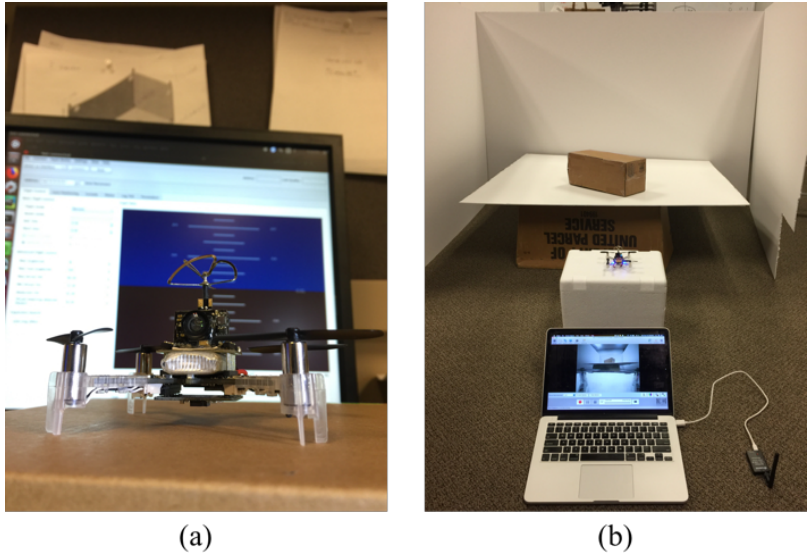

**Figure 2 (a) Aerial platform used in this study; (b) The hardware configuration and the selected inspection test case.**

In this experiment, we use the Xbox controller to manually move UAV around the target with FPV camera pointed to the target box. The onboard camera captures real-time images at 15 HZ, and those images are wirelessly transferred into the laptop through the video receiver and saved as a video file for post-processing.

**RESULTS**

To evaluate the efficiency of the proposed method, we implement the algorithm on NVIDIA Jetson TX2, a small size embedded module for robotic and AI computing. The small payload of the module make it possible to be equipped on commercially UAVs to support autonomous outdoor tasks (Shang and Shen 2018). The captured image stream was downsampled to 1 HZ and sequentially feed into this board for real-time processing. The onboard FPV camera is pre-calibrated using a 7×8 checkerboard, and the UAV pose is initialized by manually selecting five control points on the first image of the stream (Fig. 3). Due to the small size of the micro UAV, we simplify the transformation matrix from camera to UAV as an identify matrix.

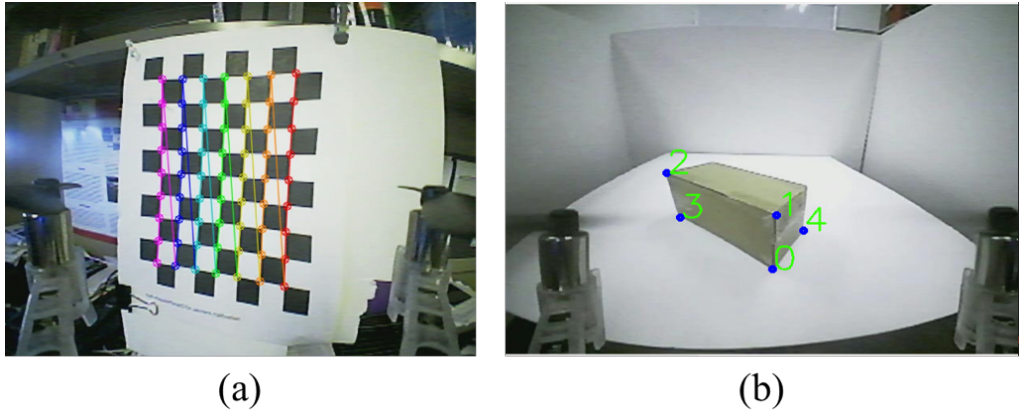

**Figure 3 (a) Camera calibration; (b) UAV pose initialization**

The tracking results at the selected frames is shown in Fig. 4. In fig. 4 (a), the LSD edges detected in the image are visualized and counted. We set the minimal length of the edges to 15 pixels that eliminates the potential outliers. In fig. 4 (b), the matching results of the model projected edges (in green) and the image detected model boundaries (in white) are presented and compared. The result shows good correspondences between the detected and the projected model. In fig. 4 (c), the updated model projection at the next frame is presented. The experiment also showed good results on continuously tracking the UAV inflight locations using the FPV camera. Fig. 5 shows the visually estimated 6d poses of the UAV during the testing period. It is computed as the inverse of the object tracking method where object is fixed at static position. The result showed that images collected with onboard camera during the aerial inspection process can be used for UAV indoor tracking and localization.

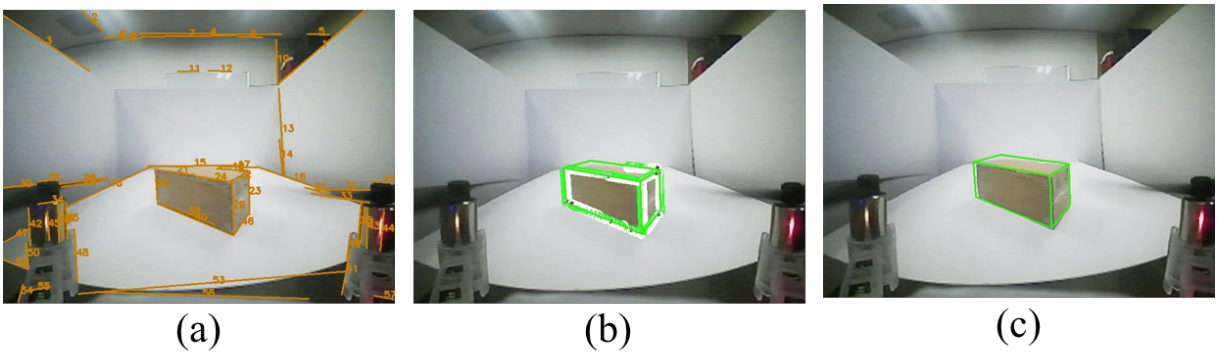

**Figure 4 (a) Edges detected in the selected frame using LSD detector; (b) Edges matching at the selected frame; (c) Projected models based on the updated pose at the next frame.**

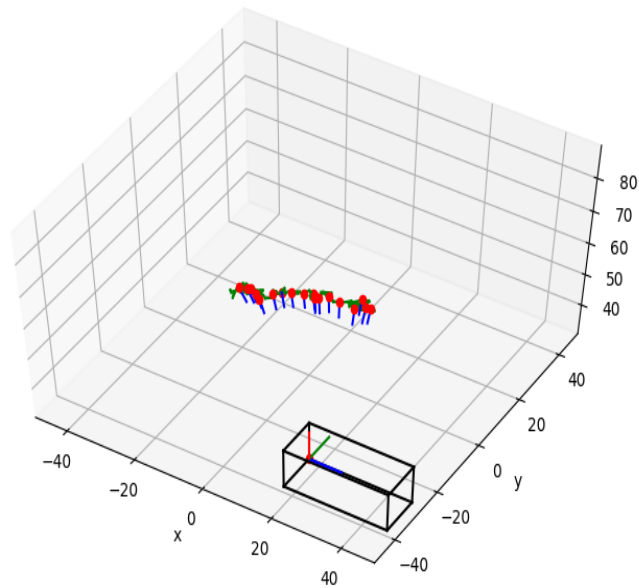

**Figure 5 Estimated relative 6d poses (in centimeters) of UAV using the images captured with onboard FPV camera during inspection**

**CONCLUSION**

In this study, a vision-model-based UAV localization and tracking method is introduced to support autonomous aerial inspection under GPS-denied environment. The proposed method is validated through a simple indoor test cases, and the result from this case showed that the image captured during aerial inspection can be simultaneously applied for UAV poses estimation. Due to the geometrical complexity of the real world structures and the significant higher environmental noise of outdoor environment, the performance of this proposed method is expected to be negatively affected, which is upon further investigations in the future.